\documentclass[letterpaper]{article} 
\usepackage{aaai2027}  
\usepackage[hyphens]{url}  
\usepackage{graphicx} 
\usepackage{natbib}  
\usepackage{caption} 
\usepackage{algorithm}
\usepackage{algorithmic}

\usepackage{newfloat}
\usepackage{listings}
\DeclareCaptionStyle{ruled}{labelfont=normalfont,labelsep=colon,strut=off} 
\floatstyle{ruled}
\newfloat{listing}{tb}{lst}{}
\floatname{listing}{Listing}

\usepackage{booktabs}

\title{
Reimagining Enterprise RAG for Multi-Agent Systems:\\
A Knowledge Layer That Organizes Before It Retrieves
}

\author{
    Garima Agrawal, Prasun Das, Akshay Sarvade, Priyanka L,\\
    Minisha N, Sravani Joshna, Hemath Manivannan, Lavanya Korada,\\
    Sambhrit Pattanayak, Srashti Gupta, Siddu R, Prasad Kalyanasundaram
}

\affiliations{
    Cognizant, Plano, Texas, USA\\
    \texttt{\{garima.agrawal, prasun.das, prasad.kalyanasundaram\}@cognizant.com}
}

\begin{document}

\maketitle

\begin{abstract}
Enterprise organizations are deploying agentic AI across the software
development lifecycle, and these agents need the behavioral logic of the
systems they act on rather than documentation written for stakeholders. Recent
work has shown that this logic can be recovered at scale: a seven-agent
reverse-engineering pipeline of our own design processed 241 repositories
across 18 services into thousands of artifacts describing scope, journeys,
business rules and acceptance criteria. Recovery, however, does not confer
usability. Business entities recur throughout such an estate under a shared
vocabulary: \textit{offer}, \textit{eligibility}, and \textit{activation} each
govern rules in several services, related in name but operationally distinct,
so an end-to-end journey implicates well over a thousand rules whose text is
near-identical across contexts that must not be conflated. Flat vector
retrieval fails here in its most consequential form, returning a structurally
identical rule from the wrong service at high similarity and with no signal of
error. Content-level ontology is the traditional remedy, but none existed, and
relating every occurrence of every recurring entity to every other would have
yielded a graph whose construction cost exceeded the engagement and whose
currency would not survive a release cycle. We place the ontology at the
\textit{artifact} level instead. Because the corpus is machine-generated, each
artifact retains its producer, component, service, domain and type, provenance
sufficient to disambiguate before any content is read. We present a knowledge
layer that \textit{organizes before it retrieves}: a self-describing catalog, a
sparse metadata registry, and a navigation hypergraph defined over artifacts
rather than extracted entities. Queries resolve to a set of artifact
identifiers to which vector search is then constrained, so the competing rule
from another service is absent from the candidate set rather than outranked
within it, and disambiguation becomes structural rather than statistical.
Constructed in under a week and admitting new services by convention, the layer
serves production SDLC agents alongside developers, quality engineers and
analysts querying it directly. We report more than 100 production defects for
which it supported root-cause analysis and resolution under developer
verification, and 35 release regression failures attributed to specific rules
and services.
\end{abstract}

\section{Introduction}
\label{sec:intro}

Enterprise software organizations are moving from using AI as a tool to
deploying it as an organizational capability. Across the software development
lifecycle, specialized agents now recover undocumented business logic, generate
tests from behavioral specifications, triage production defects against
requirements, assess merge-request risk, and attribute release failures to rule
violations. Each reasons from a distinct professional standpoint, and each
requires knowledge of the application at a level of detail that stakeholder
documentation does not supply: the behavioral logic embedded in source code,
configuration, and tests. Prior work has established that this logic can be
recovered at scale \cite{agrawal2026, agrawal2024kg}. Applying a seven-agent
reverse-engineering pipeline of our own design to the estate reported here, we
processed 241 repositories across 18 services into several thousand structured
artifacts spanning scope definitions, user journeys, business rules, acceptance
criteria, and gap registers, each finding carrying a reference to its source.
The behavioral specification of the estate, which had never been written,
thereby came into existence as structured data.

Recovery, however, does not confer usability. This paper addresses the problem
that follows it: how a corpus of machine-generated requirements artifacts must
be organized if downstream agents are to retrieve from it with the precision
their tasks demand.

\noindent\textbf{The Organization Problem.}
The estate is orchestrated through 14 business process workflows. A single
workflow may invoke as many as 112 endpoint-level requirements documents in one
execution, and an end-to-end journey such as order placement, order submission,
or customer update implicates well over a thousand business rules.
Volume alone would be manageable. The compounding difficulty is that the
estate's business vocabulary is shared across the services that compose it.
Terms such as \emph{offer}, \emph{eligibility}, \emph{activation},
\emph{order}, and \emph{provisioning} recur throughout, denoting related but
non-interchangeable concepts: \emph{offer} governs eligibility rules in the
ordering service, definition and validity rules in catalog, and discount rules
in billing. The artifacts describing them consequently exhibit substantial
textual overlap while remaining operationally distinct.

Consider what resolving a production defect requires under these conditions.
The traversal runs from workflow to endpoint document to business rule to
owning repository to change site. Depth is not the principal obstacle;
precision under vocabulary collision is. What must be retrieved is the rule
governing the service in which the failure occurred, and not its structural
twin in an adjacent service that participates in the same workflow.

\noindent\textbf{Why Retrieval Alone Is Not Enough.}
Embedding a corpus into a vector store and retrieving by cosine similarity
presumes that proximity in embedding space corresponds to operational
relevance. Vocabulary collision inverts that presumption: the highest-ranked
result is frequently drawn from the wrong operational context. What
distinguishes this failure from ordinary retrieval error is that it leaves no
trace. The returned text is well-formed and topically correct, the similarity
score is high, and the consuming agent proceeds to reason over it without any
indication that its evidence is misattributed.

Content-level ontology is the established corrective: extract entities, declare
relations, and align vocabularies across the estate
\cite{gruber1993, noy2000prompt}. Two considerations made it impracticable
here. The first is combinatorial. Each recurring entity appears in several
services and relates to every other entity it transacts with, so the graph an
ontology would have to construct and keep consistent grows with the product of
those occurrences rather than with the number of terms; across a rule
population of this size, the alignment effort would constitute the greater part
of the engagement rather than a preparatory step. The second is currency. Rules
are revised on a release cadence, so an ontology delivered after months of
specialist effort would describe a system that had already moved, while the
agents depending on it required knowledge of the system as deployed.

Graph-based retrieval \cite{edge2024graphrag, guo2024lightrag} induces an
entity graph from corpus text and traverses it at query time, with demonstrated
benefits for multi-hop reasoning. A graph induced from colliding text, however,
inherits the collision: the extractor must determine whether two occurrences of
\emph{offer} constitute one node or two, which relocates the disambiguation
problem to the construction phase rather than eliminating it.
Each of these approaches treats the corpus as opaque text whose organizing
structure must be reconstructed from content. That premise does not hold for a
machine-generated corpus.

\noindent\textbf{Organize Before You Retrieve.}
The corpus under consideration possesses a property that distinguishes it from
the document collections retrieval systems are typically designed for. Each
artifact was emitted by a known producing agent, for a known component, within
a known service and domain, under a declared type, with per-finding references
to source. The structure required for disambiguation was therefore never latent
in content. It was available at the artifact layer, in the circumstances of
production, naming, and relation, prior to any content being read.

Our response was not to forgo ontology but to relocate it to a level the
organization can sustain, imposing structure over the artifacts rather than
over the entities they contain. We term this approach \textbf{KB Cataloging},
after the apparatus by which a library renders most of its holdings irrelevant
to a given inquiry before any volume is opened: a classification scheme, a
catalog, and an index, none of which requires reading the collection to
construct.

The mechanism this yields at query time is the framework's principal
contribution. Intent classification and the catalog resolve an incoming
question to a set of artifact identifiers, and the vector query is constrained
to that set (Section~\ref{sec:design}). The competing rule from an adjacent
service does not participate in the similarity computation at all, and
disambiguation is accordingly effected by the organization of the corpus rather
than by the distribution of its scores.

\noindent\textbf{Contributions}
\begin{enumerate}
\item \textbf{Artifact-level ontology} as a design principle for enterprise
knowledge layers serving multi-agent SDLC systems, with the conditions under
which organizing artifacts substitutes for modeling their contents stated
explicitly (Section~\ref{sec:design}) and the boundary of that substitution
stated as a limitation (Section~\ref{sec:closing}).

\item A \textbf{four-component knowledge layer} comprising a self-describing
artifact catalog, a sparse metadata registry, a navigation hypergraph defined
over artifacts rather than extracted entities, and identifier-constrained
retrieval with a confidence gate and runtime context graphs. It was
constructed in under a week, admits new services by convention rather than
schema revision, and refines its own metadata from every query through
write-back.

\item \textbf{Role-aware knowledge composition}, whereby SDLC agents with
differing professional concerns, together with developers, quality engineers,
and analysts, enter a single structure at the granularity their task requires,
constituting a knowledge-serving layer rather than a search index.

\item \textbf{Deployment evidence} (Section~\ref{sec:eval}) from production
use: more than 100 defects drawn from the live defect queue for which the layer
supported root-cause analysis and resolution under owning-developer
verification, 35 release regression failures attributed to specific rules and
services, several further agent consumers, and direct querying by
practitioners.

\item An account of the \textbf{design decisions and boundary conditions} that
shaped the framework, together with the trajectory from the deployed catalog to
a fuller knowledge-serving layer (Section~\ref{sec:closing}).
\end{enumerate}

\section{Design: The KB Cataloging Framework}
\label{sec:design}

The design proceeds from a single organizing principle: every decision that can
be settled at ingestion time is settled there, so that query time incurs only
the cost of what remains. Classification, placement, relationship discovery,
and metadata capture are performed once, when an artifact enters the system.
Retrieval thereafter navigates a pre-organized structure rather than reasoning
over unstructured content.

\begin{figure*}[t]
\centering
\includegraphics[width=\textwidth]{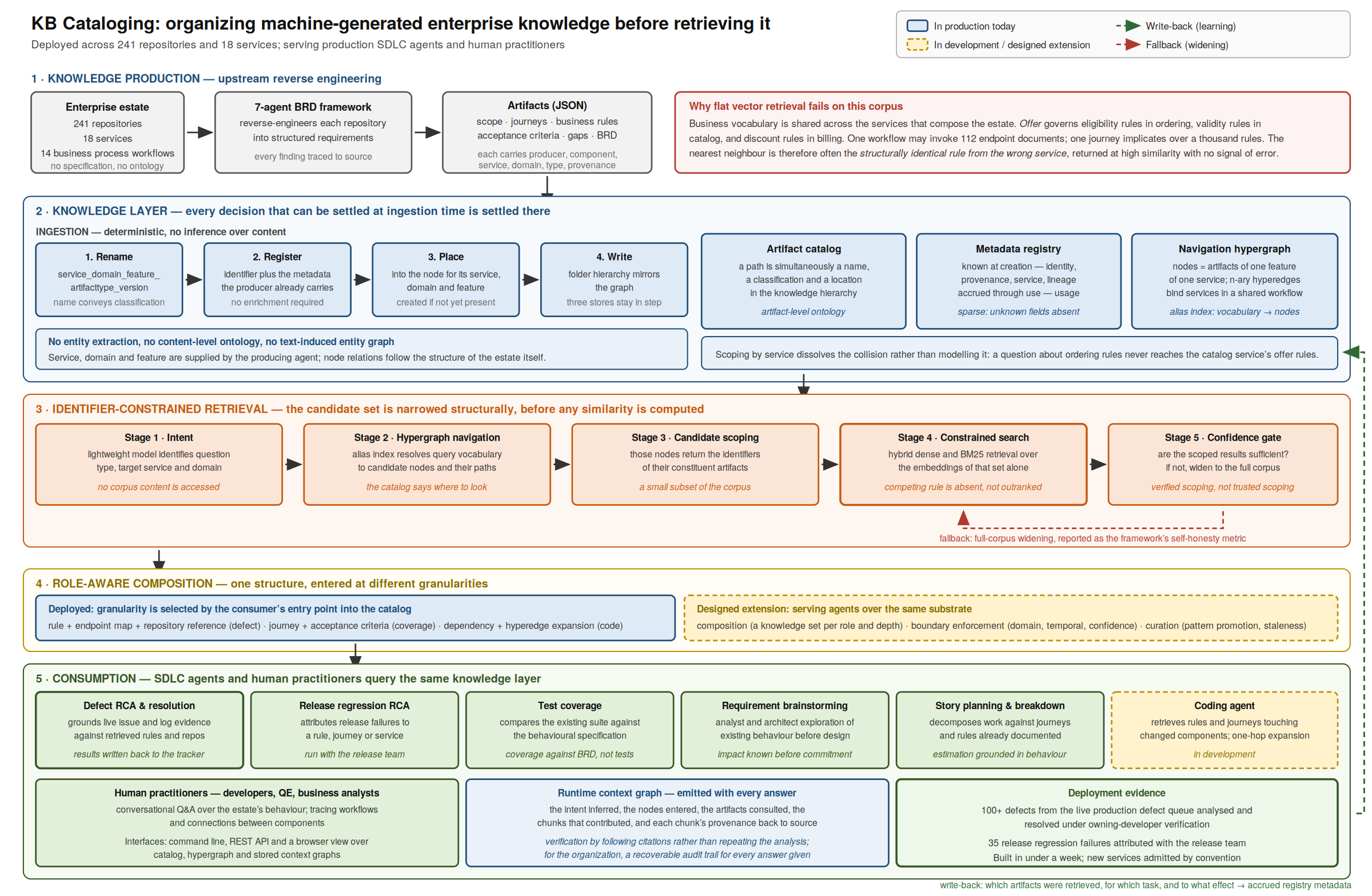}
\caption{End-to-end knowledge flow. An upstream reverse-engineering pipeline
produces artifacts; the knowledge layer organizes them at ingestion into a
catalog, registry, and navigation hypergraph; retrieval narrows to a set of
artifact identifiers before any similarity is computed; SDLC agents and human
practitioners consume the result at differing granularities. Solid borders
denote components in production; dashed borders denote components in
development and the designed extension discussed in Section~\ref{sec:closing}.
All identifiers shown are synthetic.}
\label{fig:arch}
\end{figure*}

Figure~\ref{fig:arch} traces the lifecycle of a unit of knowledge through the
system. At \textbf{ingestion}, an upstream agent emits structured artifacts as
JSON with provenance intact, and the knowledge layer renames, registers,
places, and writes each one, updating three representations in synchrony: the
artifact store, the metadata registry, and the navigation hypergraph. At
\textbf{query} time, intent classification and the hypergraph reduce a question
to a set of artifact identifiers, the vector query is constrained to that set,
and a confidence gate determines whether the scoped evidence is sufficient.
Through \textbf{write-back}, each query records which artifacts were retrieved,
for which task, and to what effect, so that the registry accrues a usage
profile rather than remaining static.

\noindent\textbf{Artifact Catalog and Placement:}
Cataloging proceeds in four deterministic steps. An artifact arriving from the
producing agent is (1)~\emph{renamed} to the convention
\path{{service}_{domain}_{feature}_{artifact_type}_v{version}.{ext}}, so that
its name alone conveys its classification; (2)~\emph{registered}, with an
identifier and the metadata its producer already carries; (3)~\emph{placed}
into the hypergraph node for its service, domain, and feature, creating that
node if it does not exist; and (4)~\emph{written} to a folder hierarchy
mirroring the graph, so that the machine-readable store, the registry, and the
navigation structure remain in correspondence.

No step requires inference over content. Service, domain, and feature are
supplied by the agent that produced the artifact, having established them in
the course of analyzing the component, and the relations between nodes follow
from the service and repository structure of the estate. This is the
framework's central economy: the classification an ontology engineer would
otherwise construct after the fact has already been performed upstream, and the
knowledge layer need only retain it rather than recover it.

\begin{table}[t]
\centering
\small
\begin{tabular}{@{}p{0.30\columnwidth}p{0.60\columnwidth}@{}}
\toprule
\multicolumn{2}{@{}l}{\textbf{Known at creation}}\\
\midrule
\texttt{artifact\_id} & unique key, never changes \\
\texttt{artifact\_type} & journey\_map, business\_rules, gap\_analysis, brd\_final, \ldots \\
\texttt{producer\_agent}, \texttt{producer\_run\_id} & which agent produced it, in which run \\
\texttt{service}, \texttt{domain}, \texttt{feature} & classification supplied by the producer \\
\texttt{repo}, \texttt{file\_path}, \texttt{version} & source repository and location \\
\texttt{confidence} & producer's confidence in the artifact \\
\texttt{derived\_from} & lineage: artifacts this was derived from \\
\midrule
\multicolumn{2}{@{}l}{\textbf{Accrued through use}}\\
\midrule
\texttt{used\_for} & task types served: defect triage, test authoring, \ldots \\
\texttt{query\_intents} & intent labels that retrieved this artifact \\
\texttt{co\_retrieved\_with} & artifacts consistently retrieved alongside \\
\texttt{usefulness\_scores} & used versus retrieved-but-ignored, per task type \\
\texttt{access\_count}, \texttt{last\_accessed} & retrieval history \\
\texttt{superseded\_by} & newer version, where one exists \\
\bottomrule
\end{tabular}
\caption{Registry schema. Fields unknown at creation are absent rather than
null, so the schema grows across services without migration.}
\label{tab:registry}
\end{table}

\noindent\textbf{Metadata Registry: Sparse, Evolving, Learning:}
The registry holds one record per artifact across two tiers, shown in
Table~\ref{tab:registry}. Fields known at creation are deterministic and arrive
with the artifact: identity, provenance, producing agent and run, service,
domain, feature, repository, version, and the lineage linking it to the
artifacts it derives from. Fields in the second tier accrue through use, and
are what render the registry a learning system rather than a static index:
which task types an artifact has served, which query intents retrieved it,
which artifacts are consistently retrieved alongside it, and how often it was
used rather than merely returned. Unknown fields are absent rather than null.

\noindent\textbf{Navigation Hypergraph:}
The hypergraph supplies the navigation intelligence between intent recognition
and vector retrieval. Its nodes group the artifacts describing a given feature
of a given service; its edges are hyperedges connecting \emph{sets} of nodes
that participate in a shared workflow. Hyperedges rather than binary edges
because the relationships are genuinely $n$-ary: an order-submission workflow
traversing ordering, catalog, and billing is one relation among three services,
not three pairwise links (Figure~\ref{fig:hypergraph}). The structure reflects
where the difficulty actually lies. Business entities recur throughout the
estate: \emph{offer} carries eligibility rules in the ordering service,
validity rules in catalog, and discount rules in billing, and relating every
occurrence of every such entity to every other is the congested graph that
content-level ontology would require us to build and maintain. Scoping by
service dissolves the problem instead of modeling it: a question about ordering
rules is answered from ordering artifacts, and the catalog service's offer
rules are never candidates. The \textbf{alias index} maps business vocabulary
onto node identifiers, resolving the terms a query uses to the nodes where the
corresponding artifacts reside.

\begin{figure}[t]
\centering
\small
\fbox{\parbox{0.93\columnwidth}{%
\textbf{Hypergraph fragment (redacted)}\\[2pt]
\rule{\linewidth}{0.4pt}\\[3pt]
\texttt{\scriptsize
NODE\ node\_S1\_ordering\_offer\_eligibility\\
\hspace*{1.2em}service S1\ \ domain ordering\\
\hspace*{1.2em}artifacts\ [journey\_map, business\_rules,\\
\hspace*{4.2em}acceptance\_criteria]\\
\hspace*{1.2em}aliases\ ["offer eligibility", "offer\\
\hspace*{3.9em}rejected", "eligible offers"]\\[3pt]
NODE\ node\_S2\_catalog\_offer\_definition\\
\hspace*{1.2em}service S2\ \ domain catalog\\
\hspace*{1.2em}aliases\ ["offer validity", "offer setup"]\\[3pt]
NODE\ node\_S3\_billing\_offer\_discount\\
\hspace*{1.2em}service S3\ \ domain billing\\[4pt]
HYPEREDGE\ edge\_order\_submission\_flow\\
\hspace*{1.2em}type\ \ \ \ \ SHARED\_FLOW\\
\hspace*{1.2em}connects\ [node\_S1\_ordering\_offer\_eligibility,\\
\hspace*{4.2em}node\_S2\_catalog\_offer\_definition,\\
\hspace*{4.2em}node\_S3\_billing\_offer\_discount]
}}}
\caption{A hypergraph fragment. The entity \emph{offer} governs rules in three
services, each operationally distinct. Nodes scope retrieval to one service;
the $n$-ary hyperedge records that a single order-submission workflow traverses
all three, which pairwise links could not express. Service identifiers are
redacted.}
\label{fig:hypergraph}
\end{figure}


\noindent\textbf{Identifier-Constrained Retrieval:}
Retrieval proceeds through five stages. \textbf{(1)~Intent classification.} A
lightweight model identifies question type, target service, and domain; no
corpus content is accessed. \textbf{(2)~Hypergraph navigation.} The alias index
resolves query vocabulary to candidate nodes and the paths among them.
\textbf{(3)~Candidate scoping.} Those nodes return the identifiers of their
constituent artifacts, ordinarily a small subset of the corpus.
\textbf{(4)~Identifier-constrained search.} Hybrid dense and BM25 retrieval is
executed over the embeddings of that set alone. \textbf{(5)~Confidence gate.}
An LLM evaluates whether the scoped results are sufficient; where they are not,
retrieval widens to the full corpus.

Stage~4 realizes the framework's central claim, and the constraint merits
precise statement. The filter is artifact identity, applied to the vector query
itself, with the consequence that the similarity computation is never presented
with the remainder of the corpus. The near-identical rule belonging to another
service is thus \emph{absent from the candidate set rather than outranked
within it}. Disambiguation follows from how the corpus was organized rather
than from how the scores happened to fall, and retrieval cost scales with the
size of the scoped set rather than that of the estate.
Figure~\ref{fig:trace} presents a worked trace.

The widening path constitutes the framework's safety property. Any scoping
mechanism may err through a misclassified intent, a missing alias, or a
misplaced artifact, and a system that trusts its scope unconditionally fails
silently. This design verifies rather than trusts, incurring the cost of a
full-corpus search where scoping proves insufficient. Section~\ref{sec:eval}
returns to the widening rate, since a framework that widens on the majority of
queries has an index that is not performing organizational work.

Every answer produces a \textbf{runtime context graph} recording the intent
inferred, the nodes entered, the artifacts consulted, the chunks that
contributed, and the provenance each chunk carries back to source. Conceived as
a debugging instrument, it proved the single most significant factor in
adoption. A developer presented with a proposed root cause is not obliged to
trust the agent's conclusion: the context graph identifies the requirements
document, the business rule, and the source reference on which it rests, so
that verification consists in following citations rather than in repeating the
analysis. For the organization, it serves as an audit trail, in that the
evidence path underlying any answer the system has produced remains
recoverable.

\begin{figure}[t]
\centering
\small
\fbox{\parbox{0.93\columnwidth}{%
\textbf{A query resolved under vocabulary collision}\\[2pt]
\rule{\linewidth}{0.4pt}\\[2pt]
\texttt{\scriptsize
QUERY\ \ "offer rejected as ineligible during\\
\hspace*{2.9em}order submission"\\[3pt]
1\ intent : defect\_rule\_lookup\\
\hspace*{2.1em}service=S1\ \ domain=ordering\\[2pt]
2\ alias "offer" -> 3 candidate nodes\\
\hspace*{2.1em}S1\_ordering\_offer\_eligibility\ \ [in]\\
\hspace*{2.1em}S2\_catalog\_offer\_definition\ \ \ [out]\\
\hspace*{2.1em}S3\_billing\_offer\_discount\ \ \ \ \ [out]\\[2pt]
3\ candidate set : 9 artifact ids\\[2pt]
4\ hybrid search over those 9 only\\
\hspace*{2.1em}R-212 offer suppressed where a prior\\
\hspace*{2.1em}\ \ \ \ \ \ order holds the line in pending state\\[2pt]
5\ confidence gate : sufficient -> answer\\[4pt]
CONTEXT GRAPH\\
\hspace*{1.2em}R-212 -> journey J-07 step 4\\
\hspace*{2.4em}\ \ \ \ -> endpoint POST /orders/submit\\
\hspace*{2.4em}\ \ \ \ -> repo <repo-ref> : OfferEligibility\\[3pt]
NOT SCORED\\
\hspace*{1.2em}R-198 (catalog) offer validity window,\\
\hspace*{1.2em}near-identical wording, different\\
\hspace*{1.2em}operational meaning. Outside the\\
\hspace*{1.2em}candidate set; never entered the search.
}}}
\caption{An illustrative step-wise retrieval trace. Service, rule, and repository
identifiers are synthetic. The final block is the framework's central property:
the competing rule never enters the similarity computation.}
\label{fig:trace}
\end{figure}

\section{Deployment and Evaluation}
\label{sec:eval}

The knowledge layer is used in two modes. Practitioners query it
conversationally, asking which rules govern a journey or how workflows connect
components across services; this conversational access is the current form of
the role-based serving described in Section~\ref{sec:closing}. SDLC agents
query it programmatically as a retrieval substrate, of which defect analysis
and release regression attribution are the two with the most operational
history. We evaluate the framework on the cost of construction and extension,
the acceptance of its output by practitioners, and retrieval accuracy, the
last of which is under evaluation at the time of writing.

The knowledge layer was built in under a week, and onboarding a service since
has required no schema revision, no ontology maintenance, and no manual
cataloging: artifacts arrive classified by the agent that produced them, and
nodes and hyperedges follow from that classification. Across the 18 services
ingested to date, none has required schema work to admit.

\noindent\textbf{Defect Root Cause and Resolution.}
The agent with the longest operational history takes a production ticket and
produces a root cause and a proposed resolution, citing the rule, repository,
and change site the fix requires. Its two generations differ in whether they
had access to the knowledge layer.
The first generation operated from issue-tracker history and runtime logs
alone. Developers found its output insufficiently concrete, because logs
indicate where behavior failed but not what the correct behavior was. Without
the requirements specification, the agent could identify a fault but not the
change required to correct it. The specification existed as the extracted
artifacts, but the agent had no means of retrieving the relevant ones.
The current agent combines evidence from the issue tracker and log platform,
accessed through the Model Context Protocol (MCP) \cite{mcp2024}, with artifacts
retrieved from the knowledge layer. More than 100 defects from the live
production queue have been analyzed during defect-fix release cycles. Each
analysis is reviewed by the developers who own the affected service, and
accepted results are written back into the issue tracker.

The framework contributes retrieval precision rather than reasoning: it scopes
retrieval to the artifacts of the service in which the failure occurred and
supplies the governing rule, the endpoint-to-rule mapping, and the repository
reference needed to localize the change. Each answer is accompanied by a
context graph identifying the artifacts and rules it draws on, which a reviewer
can check directly.
The agent's output distinguishes three outcome classes rather than a binary
actionable/not-actionable, and this distinction generalizes beyond this agent.
\emph{Resolution proposed} indicates that an error payload and a localized
failure site are both present; the output carries root cause, change site,
repository, rule references, and a verification step. \emph{Analysis required}
indicates that evidence was retrieved but a specific signal is absent, which
the output names. \emph{Evidence blocked} indicates an operational obstruction
that prevented collection. The second case is a data problem for the analyst
and the third an infrastructure problem for the platform team; conflating the
two produced most of the unusable output in the first generation.

\noindent\textbf{Other Agents Using the Knowledge Layer.}
A separate agent analyzed 35 regression failures reported during feature
releases, attributing each to a specific rule, journey, or service, and
required no modification to the knowledge layer. A test coverage agent assesses
the existing test suite against documented journeys and acceptance criteria. A
coding agent, currently in development, retrieves the rules and journeys
touching changed components and performs one-hop hyperedge expansion to
identify dependent services. Requirement brainstorming and story breakdown
agents provide analysts and architects with existing behavior before new work
is specified. Practitioner access is provided through a command-line interface,
a REST API, and a browser interface over the catalog, hypergraph, and stored
context graphs.

Each agent enters at the granularity its task requires; none required a change
to the catalog, registry, or hypergraph, and several were built after the
registry was specified.

\noindent\textbf{Retrieval Behavior.}
Latency is addressed by the design rather than measured after the fact. Because
the candidate set is narrowed before any similarity is computed, search cost
scales with the scoped set rather than the corpus, and the artifacts are JSON
and Markdown files that parse quickly. Retrieval is interactive in both
conversational and agent use, and latency has not been an operational concern
in deployment.
The property under evaluation is accuracy: whether a query returns the rule
governing the service in question rather than a near-identical rule from an
adjacent one. Two measurements are in progress. The first is the widening rate,
the fraction of queries on which the confidence gate falls back to the full
corpus, which indicates how often scoping is sufficient. The second is an
adjudicated comparison on ambiguity-weighted queries, which requires
practitioners familiar with the services to label results.

\section{Related Work}
\label{sec:related}

\noindent\textbf{Retrieval-augmented generation.}
RAG \cite{lewis2020rag, gao2023survey} is the foundation on which this work
builds rather than an alternative to it. We do not displace vector retrieval;
we prepend an organization layer that narrows the candidate set so that
similarity operates where it constitutes a reliable signal. The contribution is
upstream of the retrieval mechanism and is accordingly compatible with advances
in embedding models, rerankers, and hybrid search.

\noindent\textbf{Ontology engineering.}
The classical response to vocabulary divergence is to model the domain
explicitly \cite{gruber1993} and to align vocabularies where they diverge
\cite{noy2000prompt}. That literature established both the value of shared
conceptualization and its cost: alignment is semi-automatic at best, and the
effort scales with the number of concepts requiring disambiguation. Our
position is not that ontology is unnecessary but that its \emph{level} is a
design variable. We classify artifacts rather than the entities they contain,
which is tractable to maintain at enterprise scale and which, on a
machine-generated corpus, is supplied largely without cost by the producing
agents.

\noindent\textbf{Graph-based retrieval.}
GraphRAG \cite{edge2024graphrag} and LightRAG \cite{guo2024lightrag} construct
entity graphs from corpus text and traverse them at query time, with
demonstrated gains in multi-hop reasoning. We employ neither, and our
navigation structure is not an entity graph. A graph induced from colliding
text inherits the collision: the extractor must determine whether occurrences
of \emph{offer} in the ordering and catalog services constitute one entity or
two, which is the disambiguation problem itself. Our hypergraph is defined over
artifacts and its nodes are fixed by provenance, so the question does not
arise. Work on knowledge graphs for hallucination reduction
\cite{agrawal2024kg} is complementary; our framework produces the organized
source such approaches would query.

\noindent\textbf{Multi-agent software engineering.}
Recovering design information from existing systems is a long-standing concern
\cite{chikofsky1990}; what has changed is that agents can now perform the
recovery at estate scale. ChatDev \cite{qian2024chatdev}, MetaGPT
\cite{hong2024metagpt}, and SWE-agent \cite{yang2024sweagent} demonstrate agent
collaboration, role-based workflows, and iterative code interaction
respectively, but each presumes that specifications or issue descriptions
already exist as input. Industrial retrieval systems make the same presumption
one layer up, including FAQ-first architectures for real-time decision support
\cite{agrawal2026faq}, which operate downstream of an organized knowledge
source. Our upstream pipeline \cite{companion} recovers specifications from
undocumented systems; the present work addresses the organization problem that
arises once they exist at scale. The two are complementary: one produces the
knowledge, the other renders it usable.

\section{Limitations, Path Forward, and Conclusion}
\label{sec:closing}

\paragraph{Boundary conditions.} The approach depends on a machine-generated
corpus whose artifacts retain provenance. On heterogeneous human-authored
material, classification would require inference from content, reintroducing the
cost the design avoids; we claim no generality beyond that boundary. Within it,
retrieval precision is bounded by placement quality, since a misplaced artifact
makes its node incorrect for every query entering through it, and we do not yet
measure placement precision independently of downstream outcomes. The framework
also inherits the coverage of its upstream agents: behavior resolved only at
runtime, rules held in stored procedures or routing configuration, and
knowledge retained only in institutional memory are absent from the artifacts
and therefore from retrieval. The deployment is a single estate, and its
outcomes are adjudicated by practitioners familiar with the services in
question, which is necessary for competent judgment and a possible source of
optimism bias.
\paragraph{From catalog to knowledge serving layer.} The deployed system is
intended as the substrate for a fuller layer that adds further agents over the
same structure: a composition agent that assembles a knowledge set matched to a
consumer's role and required depth; a boundary agent that enforces domain,
temporal, and confidence separation, so that legacy and target behavior are not
combined; and a curation agent that promotes recurring observations into
reusable patterns and flags stale artifacts. Each has a limited form in the
deployed system, as service-level scoping, as the confidence gate, and as the
write-back mechanism. The sequencing is deliberate: organizing at the artifact
layer defers entity-level modeling until accumulated usage justifies it, rather
than requiring it in advance. Nearer-term work follows the same principle,
instrumenting retrieval quality per agent and re-ingesting on code change so the
layer tracks a changing codebase. The design also admits brand as a further
classification axis without structural change, should the estate extend to
multiple brands.

\paragraph{Conclusion.} When enterprise knowledge is produced by agents rather
than written by people, organizing it is a problem distinct from any single
agent's capability. Flat retrieval fails under vocabulary collision, and
content-level ontology addresses it at a cost repaid at every release. KB
Cataloging places the ontology at the artifact layer instead and constrains
retrieval to artifact identifiers, so that a rule from an adjacent service is
absent from the candidate set rather than outranked within it. Deployed on an
estate of 241 repositories across 18 services, it serves production SDLC agents
and practitioners directly, with more than 100 production defects analyzed and
resolved under developer verification. The principle is general: when a corpus
carries provenance, the structure required for precise retrieval already exists
within it, and organizing on that structure performs the disambiguation that
would otherwise require an ontology.

\bibliography{iaai27refs}


\end{document}